\documentclass[preprint,12pt,authoryear]{elsarticle}

\usepackage{amssymb}
\usepackage{amsmath}
\usepackage{array}
\usepackage{graphicx}
\usepackage{url}

\usepackage{algorithm}
\usepackage{algpseudocode}
\usepackage{algorithmicx}

\journal{Engineering Applications of Artificial Intelligence}

\begin{document}

\begin{frontmatter}



\title{Bed-Attached Vibration Sensor System: A Machine Learning Approach for Fall Detection in Nursing Homes} %% Article title


\author[idea]{Thomas Bartz-Beielstein} 
\author[iam]{Axel Wellendorf} 
\author[idea]{Noah Pütz, Jens Brandt, Alexander Hinterleitner, Richard Schulz, Richard Scholz}
\author[hsbund]{Olaf Mersmann}
\author[t4k]{Robin Knabe}

\affiliation[idea]{organization={Institute IDE+A, TH~Köln},%Department and Organization
            addressline={Steinmüllerallee~1}, 
            city={Gummersbach},
            postcode={51643}, 
            country={Germany}}
\affiliation[iam]{organization={Institut für Allgemeinen Maschinenbau, TH~Köln},%Department and Organization
            addressline={Steinmüllerallee~1}, 
            city={Gummersbach},
            postcode={51643}, 
            country={Germany}}
\affiliation[t4k]{organization={tecfor care GmbH},%Department and Organization
            addressline={Fraunhoferstra\ss{}e~8}, 
            city={Gummersbach},
            postcode={51647}, 
            country={Germany}}
\affiliation[hsbund]{organization={Federal University of Applied Administrative Sciences},%Department and Organization
            addressline={Willy-Brandt-Stra\ss{}e~1}, 
            city={Brühl},
            postcode={50321}, 
            country={Germany}}

%% Abstract
\begin{abstract}
The increasing shortage of nursing staff and the acute risk of falls in nursing homes pose significant challenges for the healthcare system. 
This study presents the development of an automated fall detection system integrated into care beds, aimed at enhancing patient safety without compromising privacy through wearables or video monitoring.
Mechanical vibrations transmitted through the bed frame are processed using a short-time Fourier transform,
enabling robust classification of distinct human fall patterns with a convolutional neural network.
Challenges pertaining to the quantity and diversity of the data are addressed,
proposing the generation of additional data with a specific emphasis on enhancing variation.
While the model shows promising results in distinguishing fall events from noise using lab data,
further testing in real-world environments is recommended for validation and improvement. Despite limited available data, the proposed system shows the potential for an accurate and rapid response to falls, mitigating health implications,
and addressing the needs of an aging population. This case study was performed as part of the ZIM Project. Further research on sensors enhanced by artificial intelligence will be continued in the ShapeFuture Project. 
\end{abstract}

\begin{graphicalabstract}
\includegraphics{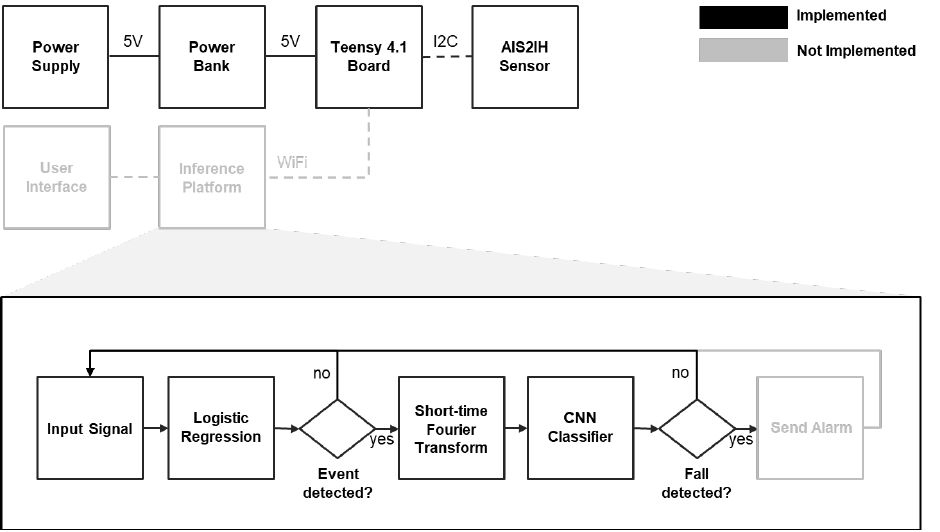}
\end{graphicalabstract}

\begin{highlights}
\item Provides a non-intrusive fall detection system integrated into care beds.
\item Illustrates the full development process from data collection to model training.
\end{highlights}

%% Keywords

\end{frontmatter}



\section{Introduction}\label{intro}

\noindent Falls in nursing homes pose substantial risks to the elderly, underscoring the critical importance of promptly identifying such incidents to mitigate immediate distress and long-term health consequences. The core objective of this research is the development of an efficient, non-intrusive fall detection system that seamlessly integrates with care beds. Extensive research from the Robert Koch Institute reveals that falls represent a significant cause of accidents, especially among the elderly, with approximately one-third of fatal accidents among those aged over 65 in the European Union attributed to falls \citep{rki24a}.
Nursing home residents, due to their vulnerability, are at an increased risk, often experiencing falls in close proximity to their beds.

It is noteworthy that nearly half of all falls result in injuries, with one in five necessitating medical treatment, as indicated by studies \citep{rubenstein-falls-2006,tine88a,sonn15a}. Of these falls, five percent result in bone fractures,
while approximately two percent lead to hip fractures. Furthermore, the prevalence of bone fractures rises significantly with age, affecting over one-third of all accidents in individuals over 70 within the EU \citep{rki24a}. The consequences of falls, especially in advanced age, are indeed severe, often leading to substantial limitations in mobility. One of the most severe injuries associated with falls is the femoral neck fracture, accounting for roughly 90\% of cases among the elderly \citep{beck11a}. Approximately half of those affected are unable to navigate stairs or engage in short walks a year after sustaining such an injury. Hip fractures, however, are even more critical, resulting in the passing of one-third of the very elderly individuals and leaving 20\% permanently dependent on care~\citep{sonn15a}. Rapid detection of such injuries can help mitigate their consequences, whereas delayed detection can result in life-threatening situations.

Also the psychological consequences of falls should be taken seriously, as they can result in decreased mobility, heightened isolation, and a notable decline in the quality of life among older individuals. In addition to the immediate physical effects of falls, they may develop a fear of falling, which, regardless of their physical condition, can compromise gait stability and lead to falls~\citep{sonn15a}. These individuals tend to restrict their mobility, decreasing their quality of life, independence, and often experiencing depression~\citep{tine88a,sonn15a,jeff14a}.

Considering these multifaceted challenges, early fall detection is of vital importance, as every minute counts in reducing suffering and long-term consequences. There is a pressing need for an alarm system capable of swiftly requesting assistance after a fall, which can alleviate the fear of falling and its associated psychological consequences. Comprehensive monitoring systems that safeguard individuals' privacy while offering timely assistance are in high demand. In the following sections, an innovative artificial intelligence (AI) based approach for the accurate detection and classification of falls occurring in proximity to beds will be implemented. This study presents a cost-effective deep learning-based fall detection system seamlessly integrated into care beds, with a focus on enhancing patient safety and ensuring privacy by avoiding the use of wearables or invasive video monitoring. It is crucial to highlight that no existing system fully satisfies these intricate requirements.

This paper is structured as follows:
Section~\ref{sec:related} provides an overview of the existing technology and approaches for fall prevention, with a particular focus on fall detection.
Section~\ref{sec:method} details the developed detection system and its constituent components. 
Section~\ref{sec:model-training} elaborates on the model training process, including data collection, augmentation, and evaluation. 
Section~\ref{sec:model-definition} presents the model architecture and the training process.
Section~\ref{sec:experiments} describes lab and real-world experiments conducted to evaluate the system's performance.
Section~\ref{sec:results} presents the results of the study and discusses the implications of the findings and potential future research directions.

\section{Related Work}\label{sec:related}

\noindent This section provides first an overview on fall occurrence and the aftermath and then gives a brief summary of the existing technology and approaches for fall prevention.

\subsection{Fall Prevention Methods}

\noindent Although various measures can be employed to prevent falls from beds, including the use of bed rails or restraints, many of these interventions are considered restrictive and can only be used under specific legal conditions, posing both ethical and safety concerns.
Bed rails, in particular, can introduce additional fall hazards and significantly impact the quality of life for those under care.

Fall-prevention measures can be categorized into three groups:
Firstly, person-centered measures which involve promoting balance, exercise for those with mobility issues, strength training, and consideration of underlying health conditions that may affect mobility. 
Secondly, medication-related measures including adjusting or discontinuing medications that may increase the risk of falls.
And lastly environment-centered measures involving proper footwear, training with assistive devices, and, critically, the elimination of tripping hazards.

While these measures can reduce the risk of falls, they cannot entirely eliminate it. Therefore, additional measures are required to detect and report falls in a timely manner.

\subsection{Fall Detection Methods}

\noindent The methods for fall-detection can mostly be divided based on the position and type of the used sensor. Body-worn sensors are embedded in clothing, accessories, and assistive devices, collectively known as ``Wearables''. User-activated or community alarm systems are based on alarm buttons that are either implemented in a wrist band or installed stationary. Automatic wearable fall detectors remove the need to press a button independently. The latter one often employs accelerometers and tilt sensors to measure the orientation of the body or shocks that occur during falls \citep{zige10a,litv08a}.

Room sensors are fixed within the living space and utilize various technologies, such as cameras, radio waves, infrared, and temperature sensors. They can monitor movement patterns and interpret falls \citep{boeh12a,kobl07a}. Especially camera-based fall detectors are highly effective \citep{loha16a}. 

Furniture-mounted sensors combine the advantages of body-worn and room-based systems. These sensors, often referred to as ``smart furniture'', can operate without the need for users to wear them, while, depending on the method, measuring necessary data to detect falls without compromising privacy. 

Current products available on the market mainly focus on detecting whether an individual is in bed or seated.
Some innovative developments aim to capture more extensive data for fall detection, such as weight shifts through the use of pressure sensors in bedposts.
Vibration based systems are installed stationary and detect falls by measuring vibrations in the floor, trying to detect the distinct pattern of a human fall \citep{alwa06a}. An extension of this approach additionally facilitates sound to further enriches the data available for the detection algorithm \citep{loha16a}.

Our study implements a vibration-based system.

\subsection{AI in Fall Detection}

\noindent The AI methods employed for fall detection are as diverse as the sensors used. Camera-based fall detection systems, for instance, utilize image recognition techniques, which represent a popular application domain for artificial neural networks \citep{foro09a,alhi14a,lu18a}. In this context, a combination of conventional methods (e.g., decision trees, $k$-nearest neighbor), more intricate machine learning approaches (e.g., support vector machine, self-organizing maps), and deep learning methods based on neural networks are utilized \citep{vall18a}.

Notably, supervised machine learning methods are frequently deployed, indicating the presence of labels for the training data, allowing for the clear differentiation between fall and non-fall data instances. Conversely, unsupervised techniques are less commonly employed, aiming to learn the normal state and classify deviations (anomalies) as potential fall events. A demonstration of this is detailed by~\citet{vinc17a}. However, for fall detection, unsupervised methods pose challenges as other events can also lead to anomalies, thereby resulting in higher false detection rates. In contrast, supervised methods are disadvantaged as they rely on labeled data, which can be difficult to obtain due to the relative rarity of fall events \citep{vall18a}.

Previous investigations into fall detection using vibrations, as relevant in this project, have employed techniques such as the Bayes classifier. This approach demonstrated relatively promising results in experiments with dummies \citep{zige10a}, albeit relying not only on vibrations but also on the additional recording of sound signals.

\subsection{Privacy in Fall Detection}

\noindent Many existing fall detection systems use sensors like cameras or microphones that could potentially infringe on users' privacy. These systems record data that can be used to monitor users in their private environments, raising significant privacy and security concerns. Data is typically exchanged over networks, possibly evaluated by third parties like cloud service providers. This situation poses notable risks from both unauthorized access and the use of data by authorized entities \citep{grun16a}. The Internet of Things (IoT) 
and Smart Home technology landscape intensifies these challenges, as many devices have weak security, handle sensitive data, and interact with technically inexperienced users \citep{kuma14a,zeng17a,psyc18a}. To address these concerns, data minimization and avoidance should be integrated into the system design from the selection of sensors to the initial data processing, aligning with privacy regulations~\citep{grun16a}.

Table~\ref{table:fall_detection_comparison} divides the currently established methods into five categories. the following four criteria were highlighted to visualize the differences and similarities between the methods, but also to illustrate the problems and advantages.
\begin{description}
  \item[Reliability] is characterized by the accurate detection of falls, the likelihood of false alarms, and resilience against human factors that may interfere with detection.problems with detection.
  \item[Privacy preservation] describes how minimally invasive the intrusion into privacy is when the corresponding system is used.
  \item[Comfort] aims at the ease of use and the effort that the user/patient needs to take to facilitate the method.
  \item[Affordability] takes the investment for the nursing home or patient into account.
\end{description}

Since user-activated or community alarm systems are dependent on the manual usage of the alarm button, depending on the fall or subsequent unconsciousness, the button might be unreachable for the patient and therefore it lacks reliability. Additionally, in case of a wearable, the patient might forget to wear it or removes it while using the bathroom, which is typically a place with a high occurrence rate of falling~\citep{litv08a}.
The later issues also apply to automatic-wearable fall detectors that additionally often produce false alarms which harms their reliability~\citep{zige10a,litv08a}. The biggest downside of highly effective camera-based fall detectors is that they are intrusive and cause privacy concerns~\citep{loha16a}. Vibration-based systems (ours) solve the problems of body-worn sensors, function independently of the patient's discipline and allow reliable fall detection~\citep{loha16a}. Combining vibration data with additional sound measurements could improve reliability~\citep{loha16a} but the downside is that installed microphones can also raise patient privacy concerns.

\begin{table}
  \centering
  \caption{Comparison of Fall Detection Systems}
  \resizebox{\textwidth}{!}{%
  \large % Increase text size
  \begin{tabular}{p{4cm}>{\centering\arraybackslash}p{3cm}>{\centering\arraybackslash}p{3cm}>{\centering\arraybackslash}p{3cm}>{\centering\arraybackslash}p{4cm}>{\centering\arraybackslash}p{4cm}}
  \hline
  Criteria & User-Activated / Community Alarm & Automatic Wearable Fall Detectors & Camera Based Fall Detectors & Vibration based systems (ours) & Vibration in combination with sound \\
  \hline
  Reliability & & & \checkmark & \checkmark & \checkmark  \\
  \hline
  Privacy Preservation & \checkmark & \checkmark & & \checkmark & \\
  \hline
  Comfort & & & \checkmark & \checkmark & \checkmark \\
  \hline
  Affordability & \checkmark & \checkmark & & & \\
  \hline
  \end{tabular}
  } % This closes the \resizebox command
  \label{table:fall_detection_comparison}
\end{table}

\section{Methods and Materials}\label{sec:method}

\begin{figure}
  \centering
  \includegraphics[width=\textwidth]{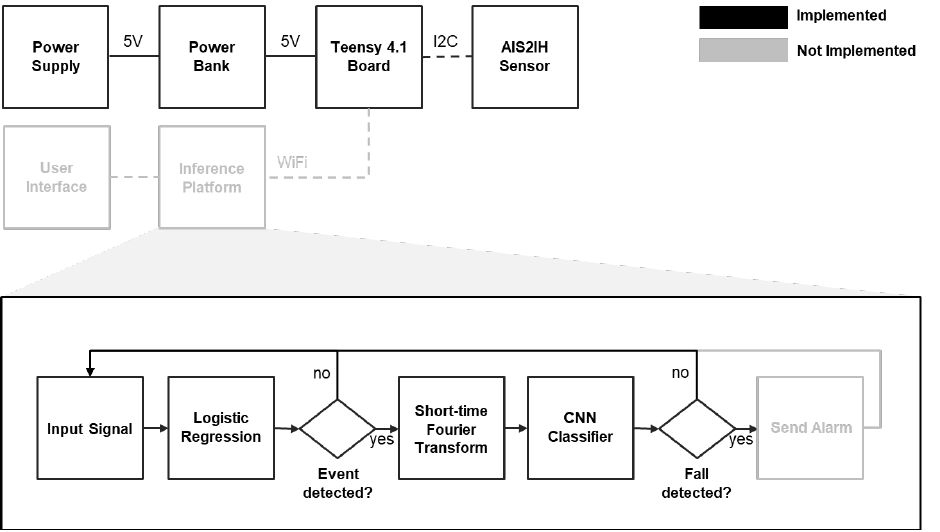}
  \caption{Overview of the developed fall detection system. The upper section illustrates the hardware components, while the lower section details the machine learning approach. The implementation of an alarm system, user interface, and central inference platform remains outside the scope of this study.}\label{fig:study-results}
\end{figure}

\noindent Figure\ref{fig:study-results} illustrates the results of the study; the muted shade of gray indicates components still to be developed for a final application.

\subsection{Developed Detection System and Component Overview}

\noindent The subsequent section describes the developed detection system and its constituent components.
The Teensy 4.1 with an ARM Cortex-M7 processor\footnote{See: \url{https://www.pjrc.com/store/teensy41.html}} is used  as the development board.
It has a processor speed of 600~MHz and 7936~KB flash storage and 1024~KB RAM \citep{mart22a}.
The Teensy 4.1 development board is selected due to its higher clock rate and larger memory compared to alternatives such as the Arduino Uno with ATMega 328 processor \citep{sih18a, hida20a}.

The developer board is connected to the adapter board of a MEMS accelerometer via an I2C bus connection.
The board stores the data measured by the MEMS acceleration sensor together with a timestamp.
Since the board lacks a separate real time clock, it must be continuously powered keep time.
Therefore, in addition to the wired power supply, the sensor module is also powered by a button cell to bridge short-term power failures. 
The collected data is stored in binary format on an integrated SD card.

During operation, the vibration measurements are intended for transmission to a central inference platform within the nursing home, aimed at facilitating fall detection through enhanced computational capabilities.
While the development of this platform and its integration with Teensy 4.1 is pending, the study encompasses the detection process.

The collected measurements are filtered using a logistic regression model, which detects possible fall events. 
For this purpose, five specific features, denoted as $f_1$ to $f_5$, are extracted and used for training the logistic regression model.
Upon the classification of the signal as a potential fall event, it undergoes a short-time Fourier transformation (STFT), yielding a spectrogram.
This spectrogram is then passed to a convolutional neural network (CNN) trained to discern human falls from other events, such as objects like water bottles.
Leveraging the logistic regression aids in significantly reducing the computational burden required for the continuous operation of the system, as it minimizes the frequency of STFTs and CNN inference runs.
An additional feature that remains pending implementation is the post-detection processing of identified human falls, envisaged to trigger a signal in a human-machine interface for the nursing staff, alongside an alarm.
So far, this integration lays a robust foundation for a cost-effective fall detection system, promising practical applications in various contexts.
The components will be explained in detail in the following sections.

\subsection{MEMS Vibration Sensor}

\noindent MEMS accelerometers can be categorized according to the different types of force measurement. The type of force measurement can be resistive or capacitive~\citep{jin21a}. Both variants have in common that the spring element of the sensor is made of silicon. The spring element functions like a bending beam that is fixed on one side and has a floating mass at the end of the free-swinging side~\citep{kazu08a}. With resistive force measurement, the displacement of the mass is determined via strain gauges. In capacitive force measurement, the mass lies between two plate capacitors and generates a change in the capacitance ratio of the two capacitors when displaced. Capacitive MEMS sensors are small, cheap, and suitable for detecting low-frequency ground vibrations~\citep{niu18a,baba21a}. Therefore, MEMS sensors based on the capacitive principle are used in the developed system.

\begin{figure}[t]
  \centering
  \includegraphics[width=\textwidth]{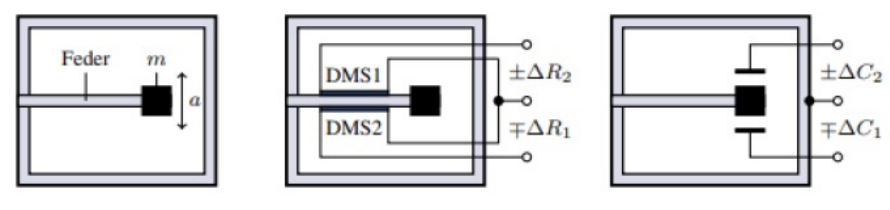}
  \caption{Structure of MEMS Sensors with Spring Element Based on the Resistive Principle (Middle) and the Capacitive Principle (Right).}\label{fig:mems}
\end{figure}

Preliminary investigations have shown that a sampling rate of 1600~Hz is necessary for precise fall classification.
The capacitive MEMS sensor selected for the planned measurement system is of the type AIS2IH\footnote{\url{https://www.st.com/resource/en/datasheet/ais2ih.pdf}} \citep{zaul23a}.
It has a sufficiently high sampling rate of 1600~Hz in all three dimensions compared to alternative sensors such as the BMA400\footnote{\url{https://www.mikroe.com/accel-5-click}} with 800~Hz~\citep{tjon21a}. 
The AIS2IH MEMS sensor is integrated into a STEVAL-MKI218V1 adapter board, which is connected to the Teensy 4.1 developer board via an I2C communication bus.

\subsection{Logistic Regression}\label{sec:logistic}

\noindent In the studied use case of human fall detection, Logistic regression serves as a primary mechanism to estimate the probability that a particular input data point belongs to one of two distinct categories: an event or no event. In this state of the detection, an event is anything that falls to the ground.

The data batches, each of 10 seconds in length, are manipulated by initially undergoing a zero-mean transformation and squaring, followed by the calculation of five pivotal features: maximum, median, mean, and 25th and 75th quantiles. 
The logistic regression model, mathematically represented as

\begin{equation*}
    P(y=1|x) = \frac{1}{1+e^{-\beta x}}
\end{equation*}

\noindent involves optimizing the components $\beta_0, \beta_1, \ldots, \beta_p$ of the vector $\beta$  to best predict the probability of an event occurrence, $P(Y=1|X)$, given $p+1$ input features  $X_0, X_1, \ldots, X_p$. 
The optimization is typically achieved through a gradient descent on a suitable loss function, often the cross-entropy loss. 
The decision about the predicted label for an observation is then made by setting a threshold, commonly 0.5; if $P(Y=1|X)$ is above this threshold, the observation is classified as an event, otherwise, it is not.
In the application described, ensuring that the logistic regression model is accurately trained and validated is of vital importance, as misclassification, especially false negatives, could result in serious events such as a person falling not being detected. 
Therefore, robust implementation and evaluation of the logistic regression model are critical to reliably filter and subsequently analyze true events \citep{mccu19a}.

\subsection{Short-Time Fourier Transform}

\noindent The STFT is a key method in signal processing that offers a way to analyze sound or vibration data in the time-frequency domain~\citep{shuv17a}. It achieves this by dividing a waveform into smaller time segments or windows and then applying Fourier transforms to each of these windows. This approach enables us to examine how the signal's frequency characteristics change as time progresses. The given signal, denoted as $x$, can be transformed into a discrete STFT representation, labeled as $F$. This transformation involves summing over all possible values of $i$, with $i$ extending from negative infinity to positive infinity. In this summation, $x(i)$ is multiplied by a window function $\omega$ centered at the time index $n$, and also by a complex exponential term $e^{-j\omega n}$:

\begin{equation*}
F(n, \omega) = \sum_{i=-\infty}^{\infty} x(i) \omega(n-i) e^{-j\omega n}.
\end{equation*}

\noindent The resulting STFT representation, $F(n, \omega)$, is informative about various aspects of the waveform, such as the frequencies present in the signal and their respective amplitudes. Taking the squared magnitude of this windowed STFT representation, denoted as $|F(n, \omega)|^2$, and concatenating it over time, a spectrogram is obtained~\citep{shuv17a}.

\subsection{Convolutional Neural Network}

\noindent CNNs represent a category of deep neural networks that are commonly employed in image classification and analysis. CNNs facilitate end-to-end learning for categorization and feature extraction. The fusion of CNNs with STFT-generated spectrograms represents a promising approach to audio or vibration classification~\citep{shuv17a,sejd09a,grun16a}. CNNs serve as feature extractors, discerning spectro-temporal patterns directly from STFT data and eliminating the need for traditional feature representations like Mel-Frequency Cepstral Coefficients (MFCCs). Advantages of this approach include simplicity and the potential for the network to learn complex and abstract concepts and features that might not be obvious to a human analyst. Nonetheless, this methodology comes with computational and memory demands, especially when processing high-resolution spectrograms~\citep{demi19a}. Careful resource allocation is vital for real-time applications. Additionally, the risk of CNNs learning unwanted patterns, such as channel distortions, necessitates diligent preprocessing and regularization.

The primary function of the convolutional layer is to establish local connections between features from the preceding layer and map their characteristics to a feature map.
This mapping is achieved through the convolution operation, where the input $I$ is convolved with the filter $F$:

\begin{equation*}
  (I*F)_{n,m}= \sum_{k=-a_1}^{a_1} \sum_{l=-a_2}^{a_2} F_{k,l} I_{n-k,m-l}.
\end{equation*}

\noindent The filter $F$ can be represented as a matrix, i.e.,

\begin{equation*}
\begin{pmatrix}
  F_{-a_1, -a_2} & \cdots & F_{-a_1, a_2} \\
  \vdots & \ddots & \vdots \\
  F_{a_1, -a_2} & \cdots & F_{a_1, a_2}
\end{pmatrix}.
\end{equation*}

\noindent To introduce non-linearity to the feature map generated by the convolution operation, a non-linear activation function is applied~\citep{beng97a}. To achieve a compression of features in connection with CNNs, max-pooling layers are often used. The max-pooling layer's purpose is to identify semantically relevant features from the previous layer. It achieves this by down-sampling the previous layer, dividing it into rectangular pooling regions, and computing the maximum value within each region~\citep{demi19a}. In summary, CNNs ability to automatically extract complex features, while simplifying the feature engineering process, makes them a compelling choice for waveform analysis. The convolutional layer, non-linear activation, and max-pooling enhance the efficacy of this approach in extracting and classifying meaningful spectro-temporal information.

\subsection{Metrics}\label{sec:metrics}

\noindent To evaluate the performance of the trained models, mainly the Recall and Precision were facilitated. They are calculated as follows:

\begin{equation*}
  \text{Recall} = \frac{\text{TruePositive}}{\text{TruePositive} + \text{FalseNegative}}
\end{equation*}
  
\begin{equation*}
  \text{Precision} = \frac{\text{TruePositive}}{\text{TruePositive} + \text{FalsePositive}}
\end{equation*}

\noindent Recall measures the model's ability to correctly identify all positive instances within the test set, making it particularly important in scenarios where missing positive cases can have serious consequences. For fall detection in nursing homes, a high Recall ensures that the system does not overlook critical fall events and was therefore of highest priority for the evaluation of the models. Precision quantifies the model's ability to correctly classify positive instances out of all instances predicted as positive, making it valuable in applications where false alarms are undesirable~\citep{power20a}. In the context of fall detection systems in nursing homes, high Precision means fewer false alarms, reducing the workload on healthcare staff. The objective was to attain optimal fall detection (Recall = 1.0) while minimizing false alarms (maximizing Precision). During evaluation, the threshold was dynamically adapted to achieve a Recall of 1.0 and then aimed to maximize Precision. The use of accuracy as a metric was not suitable because the dataset exhibited a significant imbalance. In Sec.~\ref{sec:augmentation}, the use of data augmentation methods to address this imbalance in training will be discussed in more detail.

\subsection{Data Augmentation}\label{sec:augmentation}

\noindent To train the classification models, vibration measurements of human falls are necessary. To avoid injuring people, crash test dummies were used for the artificial generation of these incidents. However, this kind of data generation is substantial effort and can hardly be automated. Data augmentation is a feasible method to generate additional data based on the dummy trials. Augmenting data can involve traditional approaches like altering or permuting the data. In addition, the introduction of noise plays an important role. Modern techniques generate new, synthetic data based on information from existing data. These methods include variational autoencoders (VAE) and generative adversarial networks (GANs), which are algorithms based on neural networks. This study examines two classical augmentation methods, namely oversampling and amplification. Further research might delve deeper into more sophisticated augmentation methods.

Oversampling is the most trivial augmentation method and simply duplicates the dummy fall samples as often as necessary. This augmentation method serves as a baseline and allows to assess the benefits of more sophisticated augmentation methods~\citep{shor19a}. Amplification is an augmentation method that might be of high relevance for the studied problem, since during data acquisition, a single dummy was utilized, and fall data exclusively reflects this dummy's weight. Amplification and damping of the signal are assumed to simulate falls of individuals with varying weights, thereby broadening the operating range of the algorithm. The amplification process initiates with the extraction of the mean and standard deviation of the signal's noise, utilizing a statistical tool called three-sigma rule. This rule states that samples of normally distributed data that are situated outside the mean plus or minus three times the standard deviation are define as outliers~\citep{lehm13a}. Within this study, this method serves to identify samples associated with specific events (samples outside of the boundary range) or noise (samples inside the boundary range) in the dataset. To center the signal around zero and prevent a zero offset for the noise, the computed mean of the identified noise is subtracted from the original signal. Subsequently, a new limit range is defined by adding and subtracting three times the noise's standard deviation from zero. Values exceeding these limits are identified as indicative of fall events. The amplification factor is then randomly selected from a uniform distribution within predetermined boundaries. Signals below the limit range are scaled down, while those surpassing it are scaled up. Finally, the signal scaling is reverted. This method facilitates targeted signal amplification exclusively for values relevant to fall events.

\begin{figure}[t]
  \centering
  \includegraphics[width=\textwidth]{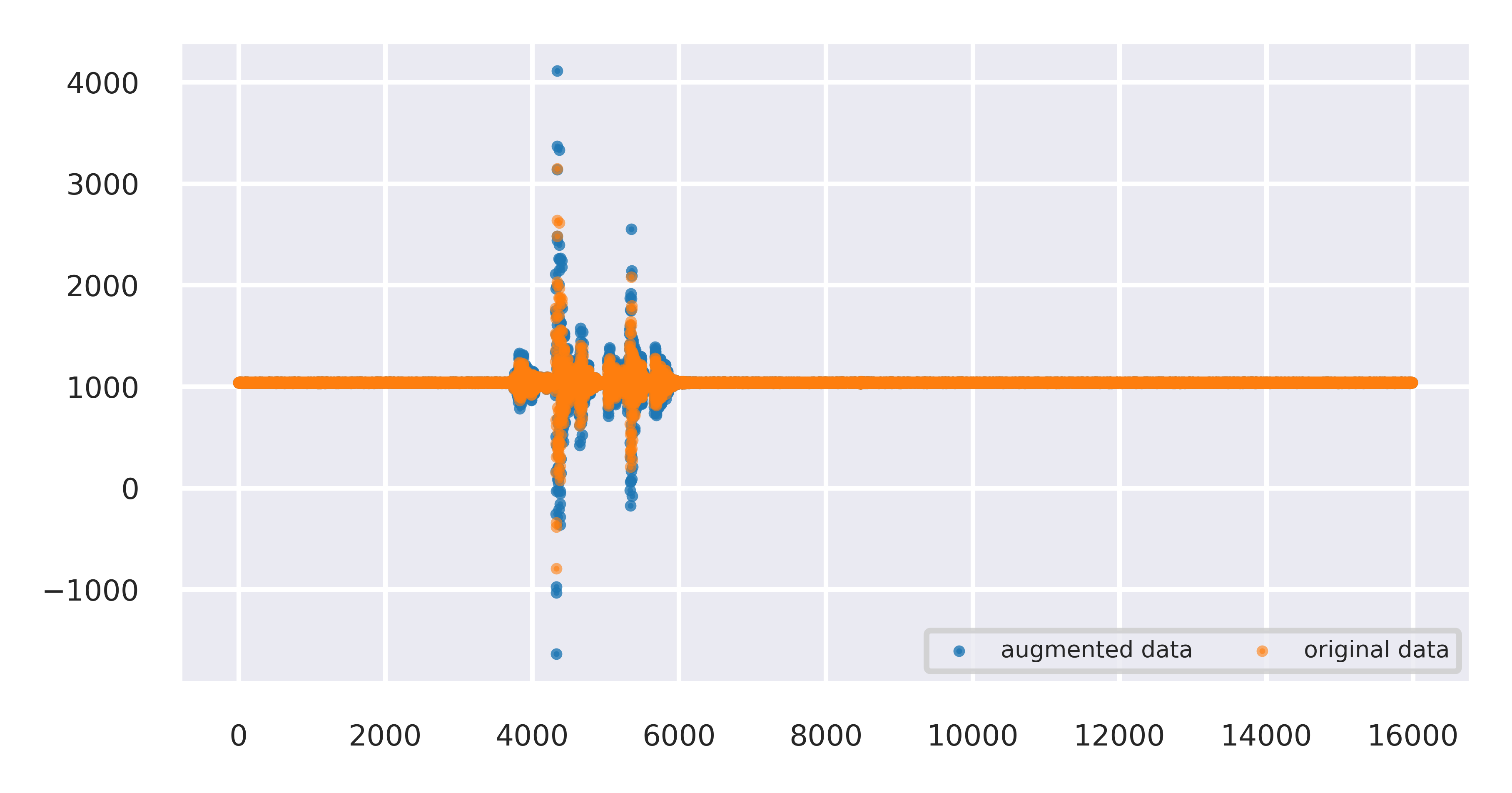}
  \caption{Positive amplification of a dummy event. The \emph{orange} scatter plot shows the original data, and the \emph{blue} plot represents the amplified signal. The x-axis indicates time steps, and the y-axis represents acceleration.}\label{fig:dummy-event}
\end{figure}

Figure~\ref{fig:dummy-event} shows an example of the positive amplification of a dummy event. The orange scatters plot shows the original data, while the blue plot shows the amplified signal. It can be seen that the amplification is only carried out at points that can be assigned to an event. Noise within the data window is not amplified, as this would distort the characteristics of a fall event.

\section{Model Training}\label{sec:model-training}

\subsection{Data Collection}

\noindent Data collection is important in the development of a sensor-based fall detection system that is able to accurately detect fall events and distinguish them from other events~\citep{loha16a}. 
Its primary purpose is to gather the necessary data for training and evaluating the fall detection system.

\subsubsection{Sensor Placement}

\begin{figure}[t]
  \centering
  \includegraphics[width=\textwidth]{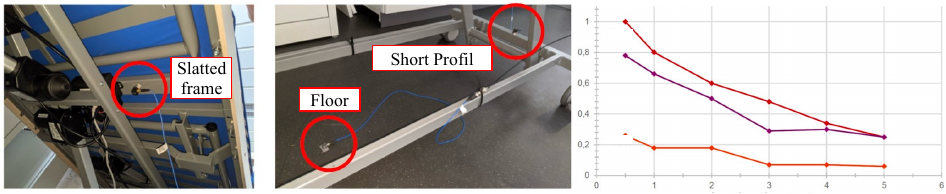}
  \caption{Lab data acquisition setup. The \emph{Left} and \emph{middle} images show sensor placements, while the \emph{right} figure displays the corresponding acceleration curves. The sensor on the slatted frame exhibits the smallest acceleration, indicating the lowest sensitivity.}\label{fig:lab-data-2}
\end{figure}

\noindent The placement of the sensors is an essential aspect of the data collection process~\citep{litv08a,loha16a,alwa06a}.
The measuring system will be integrated into the care bed in the final application. 
To analyze the data quality depending on the measuring point, the sensor data is recorded at three different positions, see Figure~\ref{fig:lab-data-2} a) and Figure~\ref{fig:lab-data-2} b). For this purpose, a 4.7~kg test specimen is dropped from a height of 1 m at different distances from the care bed. The comparison of the maximum measured amplitudes as a function of the distance between the sensor and the impact enables the optimum placement to be determined. Measuring point I, on the upper bed frame, has the smallest maximum acceleration of 0.25 g. The acceleration of point II and point III follow following a similar curve, although the maximum acceleration at positions II is 1.01 g and III is 0.78 g.
Due to the low acceleration and the associated low signal strength at measuring point I, for further tests the sensors are mounted at positions II and III.

\subsubsection{Event Generation}

\noindent To create a comprehensive data set a variety of different floor vibrations has to be measured. 
Beside a dummy fall test, different events have to be simulated, for example, falls with everyday objects such as bottles, dumbbells and personal care items, as well as dynamic actions such as jumping, tipping of chairs and furniture movements. 
The events are carefully selected to cover different scenarios and potential fall situations.

\begin{table}[h!]
  \centering
  \caption{Experimental Setup Overview}\label{table:lab-data-1}
  \begin{tabular}{|c|c|c|c|c|}
  \hline
  Setting & Floor & Storey & Distance to sensor [m] & Room Size \\
  \hline
  1 & PVC & Ground & 1 & large \\
  2 & carpet & Ground & 1 & large \\
  3 & PVC & Upper & 1 & small \\
  4 & carpet & Upper & 1 & small \\
  5 & PVC & Ground & 3 & large \\
  6 & carpet & Ground & 3 & large \\
  7 & PVC & Upper & 3 & small \\
  8 & carpet & Upper & 3 & small \\
  \hline
  \end{tabular}
  \end{table}
  
  \begin{table}[h!]
  \centering
  \begin{minipage}{0.48\textwidth}
  \centering
  \caption{Dropped Objects}\label{table:lab-data-3}
  \begin{tabular}{|c|l|}
  \hline
  Event ID & Event Description \\
  \hline
  1 & NL Bottle \\
  2 & Dumbbell \\
  3 & Personal items \\
  4 & Suitcase \\
  5 & Mobile phone \\
  6 & Crutches \\
  7 & Water crate \\
  8 & Two books \\
  9 & Moving furniture \\
  10 & Tray with cutlery \\
  11 & Dummy \\
  12 & Keychain \\
  13 & Jumping \\
  14 & Beverage bottle \\
  15 & Tipping chair \\
  \hline
  \end{tabular}
  \end{minipage}
  \hfill
  \begin{minipage}{0.48\textwidth}
  \centering
  \includegraphics[width=\textwidth]{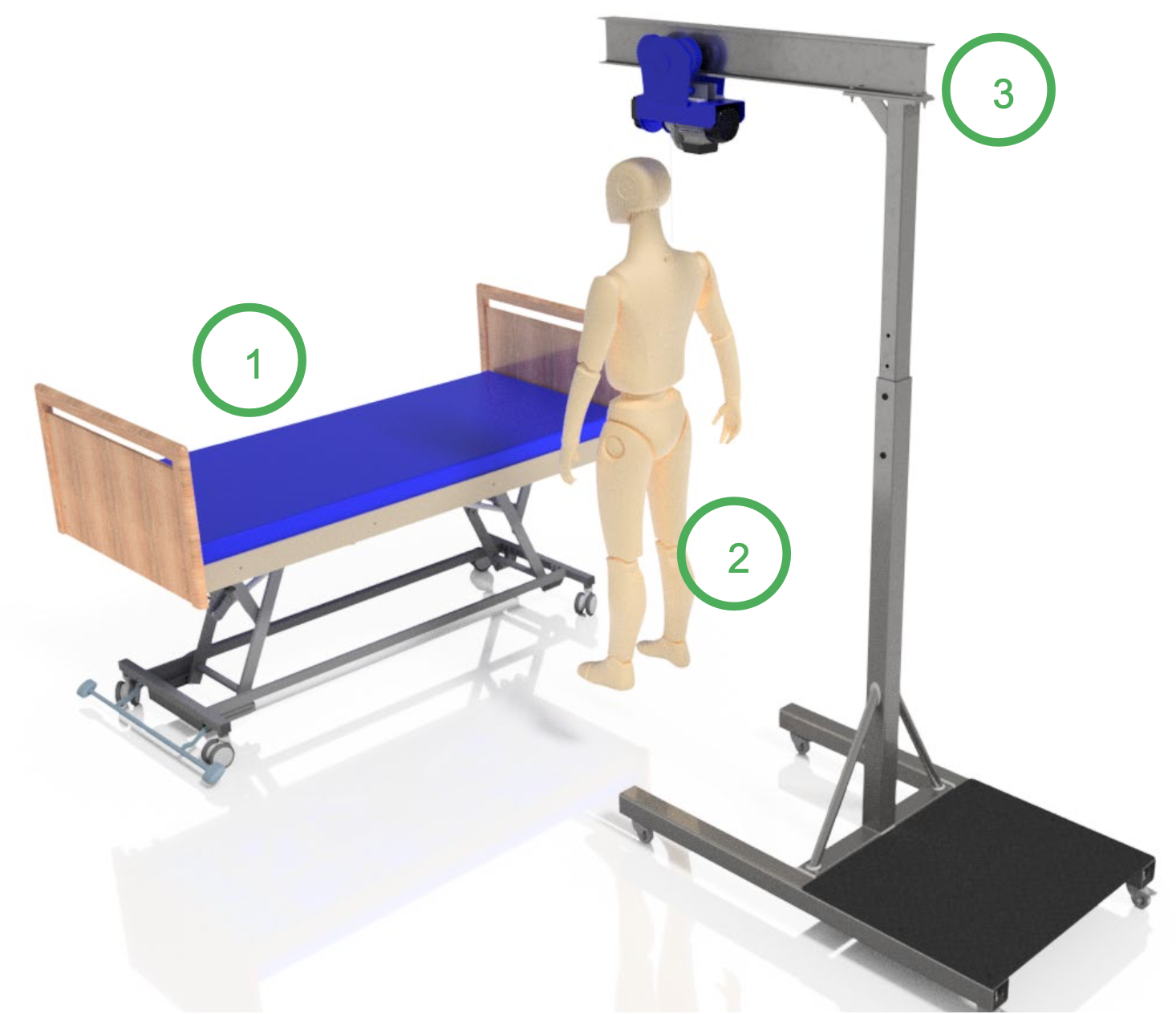}
  \caption{Visualization of the test setup, where a crane lifts the dummy to simulate a fall near a bed equipped with sensors for data measurement.}\label{fig:dummy-setup}     
  \end{minipage}
\end{table}

Table~\ref{table:lab-data-3} shows the fall events included in the data set. In addition, each event is measured under different configurations of room size, floor level, floor material and distance of the sensor to the fall. All variants are listed in  of Table~\ref{table:lab-data-1}. The test setup is designed to mimic real-life conditions as closely as possible to ensure the accuracy and relevance of the data collected. The setup is visualized in Fig~\ref{fig:dummy-setup}. To ensure repeatable tests, a device is built that allows the dummy to fall in a defined way. The most important parameters of the dummy setup are:

\begin{enumerate}
\item Maximum distance between the dummy and the floor: approx. 350~mm
\item Triggering of the drop test with a quick-release lever
\item Test dummy: I.A.F.F. Rescue Randy made of PVC, weighing 75~kg and 1.83~m tall
\end{enumerate}

This setup enabled the systematic and controlled generation of fall events and ensures that data collection is not only comprehensive but also consistent. The data set for fall detection is created on the basis of more than 1000 fall events. Each event is measured three times with three sensors and eight settings each. In addition, around 20,000 negative events are extracted from the 24-hour measurements to strengthen the data diversity with recordings without falls. The dummy's posture is varied, resulting in different fall patterns. This variation contributes to the diversity of the dataset and allows the system to recognize and adapt to different realistic fall scenarios. This is beneficial for training AI models as it adds an additional layer of complexity and realism to the dataset~\citep{zige10a,litv08a}.

\subsubsection{Field Tests in Nursing Homes}

\noindent To ensure that the AI models can also be trained with real falls, field tests are carried out over a period of 6 months in a retirement and nursing home.

The sensor unit described in Sec.~\ref{sec:method} is manufactured 20 times for the field tests. The field test study is carried out in 10 rooms of the retirement and nursing home. Two sensor units are installed in each room at positions II and III. The rooms differ in size (small and large), number of residents (1 and 2) and floor (ground floor, first floor and second floor). The floor type (PVC) and the type of bed used remain the same. The measurement data is read out at intervals of 4 weeks and stored anonymously. Over the duration of the field tests, a total of 29 falls occurred in the rooms equipped with sensor units. During readout, the sensors were found to be disconnected from their power supply a total of 33 times. As a result, some sensor units were unable to record any or only partial data within a readout interval. This means that not all falls can be evaluated.

\section{Model Definition and Tuning Results}\label{sec:model-definition}

The vibration signal employed for fall detection was sampled at a frequency of 1600~Hz. Utilizing a shifting window approach, the detection algorithm received 10-second segments of this signal for classification. To serve as a pre-filter, the logistic regression was implemented using the scikit-learn library. As explained in Sec.~\ref{sec:logistic}, the extraction of several features from the vibration signal batches was conducted. Given these features, the logistic regression was trained to differentiate between fall events, not only limited to human falls, and the absence of any event. 

\begin{figure}[t]
  \centering
  \includegraphics[width=\textwidth]{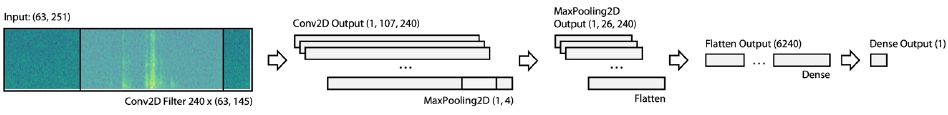}
  \caption{Structure of the CNN. The input has a shape of (63, 251) pixels, and 240 Conv2D filters of size (63, 145) are applied. The resulting output is reduced in dimension using a MaxPooling2D layer (1, 4), flattened, and then passed to the output layer.}\label{fig:cnn}
\end{figure}

The CNN implementation was carried out using Keras. The original signal batches underwent a STFT, generating spectrograms utilized for training the CNN. The architecture includes a convolutional layer with ReLU (R(z)=max(0,z)) activation function, no padding, and strides set to one. Subsequently, a max-pooling layer and a flatten layer were incorporated, followed by a fully connected output layer with a sigmoid activation, as depicted in Fig.~\ref{fig:cnn}. 

The CNN's hyperparameters were optimized using the Hyperband tuner integrated in Keras. The method is based on the successive halving algorithm and allows hyperparameter tuning by testing several settings in parallel and sorting out bad runs at an early stage~\citep{li18a}

The optimized features included the number of filters, kernel size width, loss function, and learning rate of the optimizer. Table~\ref{tab:hpt} provides an overview of the hyperparameter search space. The goal of the tuning process was to optimize precision on the validation dataset while maintaining a recall value of 1. This involves setting a classification threshold that ensures all relevant instances are captured (recall of 1) while maximizing the precision of the predictions. Consequently, the hyperparameter tuner aimed to identify a configuration that accurately classifies each fall while minimizing false alarms. The tuning process was conducted for a maximum of 20 epochs per run and the validation loss as the stopping criterion. The outcomes of the tuning process are presented in Table~\ref{tab:hpt}.

\begin{table}[t]
  \caption{Hyperparameter Tuning Search Space}\label{tab:hpt}
  \centering
  \resizebox{\textwidth}{!}{%
    \begin{tabular}{>{\raggedright\arraybackslash}p{4cm}>{\raggedright\arraybackslash}p{8cm}>{\centering\arraybackslash}p{4cm}}
      \hline
      \textbf{Parameter} & \textbf{Search Space} & \textbf{Result} \\
      \hline
      Number of Filters & $8, \ldots, 256$ (step=8) & 240 \\
      Kernel Size Width & $5, \ldots, 200$ (step=5) & 145 \\
      Loss Function & Binary Cross-entropy, Binary Focal Cross-entropy, Sigmoid Focal Loss Entropy & Binary Cross-entropy \\
      Learning Rate & $0.1, 0.01, 0.001$ & 0.01 \\
      \hline
    \end{tabular}
  } 
\end{table}

Final training was performed using the tuned settings with a stratified $k$-fold cross validation with $k$ set to 5.
Additionally, as explained in Sec.~\ref{sec:augmentation}, oversampling and amplification augmentation were facilitated to enrich the dataset with additional fall events, thereby addressing the imbalance in the initial dataset. 
The final training was carried out for 75 epochs, employing the Adam optimizer. 

\section{Experiments}\label{sec:experiments}
\subsection{Laboratory Experiments}\label{sec:lab-experiments}
\noindent In the pursuit of determining the optimal model architecture for the use case, a structured experiment pipeline was established. 
This pipeline was essential in evaluating various experiments while ensuring comparability for retrospective analysis.
Central to our methodology was the implementation of stratified $k$-fold cross-validation due to the imbalanced dataset to validate experiment results.
The experiments focus on the second phase of the two-step procedure, which is responsible for classifying different event types; the first phase uses logistic regression to differentiate event data from noise.

\begin{algorithm}[H]
  \caption{Experiment Pipeline}
  \label{alg:pipeline}
  \begin{algorithmic}[1]
    \State $\text{Dataset} \gets D$ \Comment{Original data}
    
    \For{$i \gets 1$ to $K$} \Comment{Stratified $k$-fold cross-validation}
      \State $D_{\text{train}}, D_{\text{val}} \gets \text{Split}(D, i)$ \Comment{Split into train and validation}
      
      \For{$j$ in AugmentationMethods} \Comment{Apply augmentation}
        \State $D_{\text{aug}} \gets \text{Augment}(D_{\text{train}}, j)$
        
        \For{$k$ in ModelTypes} \Comment{Train models}
          \If{$k$ is Baseline}
            \State Train $M_{\text{baseline}}$ on $D_{\text{train}}$
          \ElsIf{$k$ is All-Inclusive}
            \State Combine $D_{\text{train}}$ and $D_{\text{aug}}$ into $D_{\text{all}}$
            \State Train $M_{\text{all-inclusive}}$ on $D_{\text{all}}$
          \ElsIf{$k$ is Augmented}
            \State Train $M_{\text{augmented}}$ on $D_{\text{aug}}$
          \ElsIf{$k$ is Two-Step}
            \State Train $M_{\text{step1}}$ on $D_{\text{aug}}$
            \State Retrain $M_{\text{step2}}$ on $D_{\text{train}}$ using $M_{\text{step1}}$ as initialization
          \EndIf
        \EndFor
      \EndFor
    \EndFor
  \end{algorithmic}
\end{algorithm}

To validate different CNN models, a specific evaluation function was employed. 
When dealing with data augmentation, it was vital to integrate this process into the framework of stratified $k$-fold cross-validation.
To identify an appropriate augmentation function, it was necessary to derive it solely from the training set.
Failing to do so could result in data leakage, even when using augmentation methods that significantly alter the original data structure. 
Despite substantial changes, the fundamental basis and underlying truth of the data persist. As a consequence, this approach resulted in a computationally intensive implementation of the $k$-fold method (see Algorithm.~\ref{alg:pipeline}). 

Every data augmentation method was evaluated on four unique models:
\begin{enumerate}
  \item a baseline model trained on original data,
  \item an all-inclusive model trained on both original and augmented data,
  \item a model exclusively trained on augmented data, and
  \item a two-step model initially trained on augmented data then fine-tuned on original data.
\end{enumerate}
These models were assessed across various settings and with different data augmentation methods. As stated in Sec.\ref{sec:metrics}, the precision of the predictions were maximized while maintaining a recall of 1.0. 

\subsection{Deployment in Nursing Home}

\noindent In the simulation of the application scenario in a nursing home, the focus was on simulating the actual operating conditions and the feasibility of the fall detection system. The data was collected with the sensor setup described in Sec.~\ref{sec:method}.
The difference to the later planned implementation is that the data has so far been stored and then analyzed and not in real time, as is later planned for the real application. With the help of these authentic vibration measurements, the accuracy of the simulation and the correspondence with real signal characteristics and subtleties could be tested.

\begin{figure}[t]
  \centering
  \includegraphics[width=\textwidth]{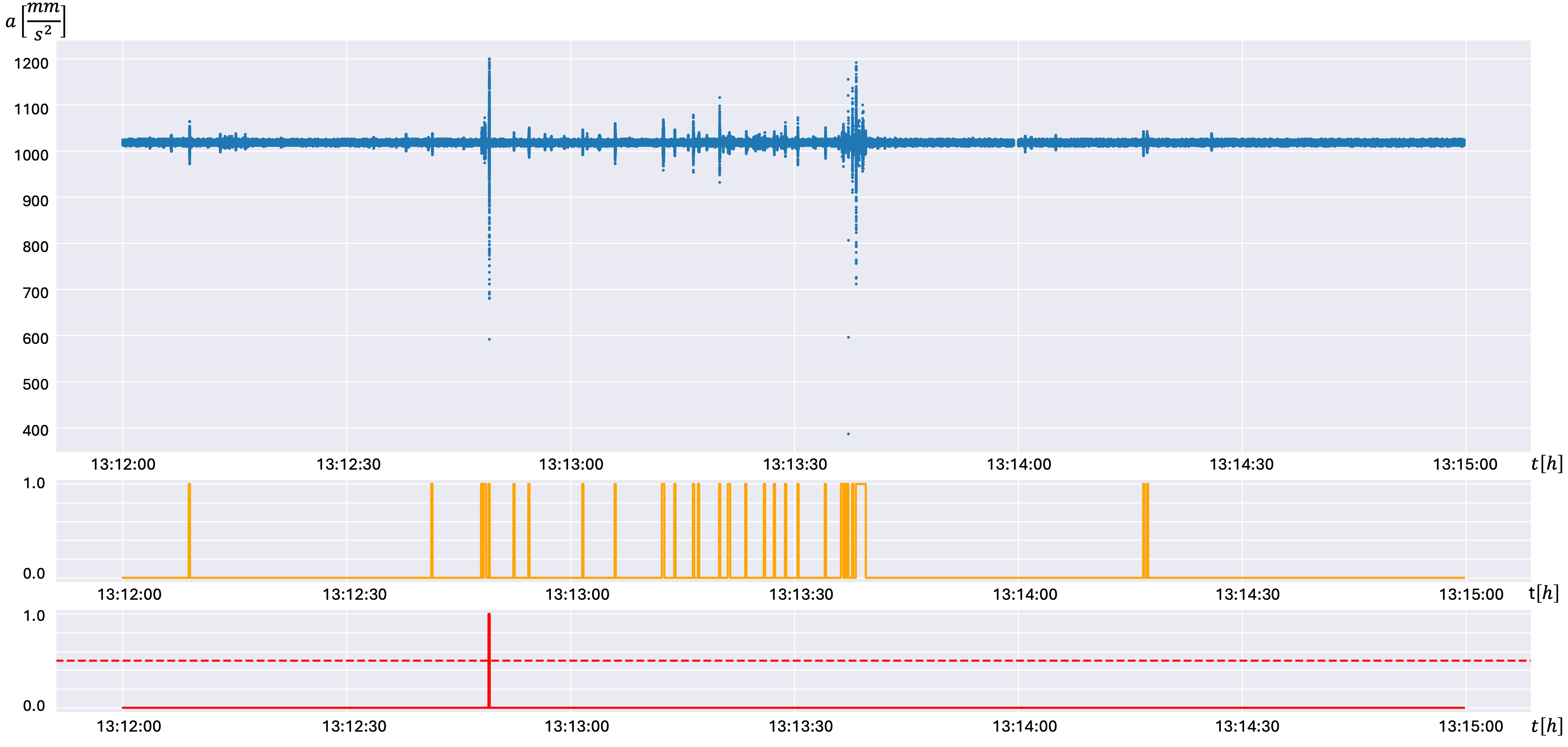}
  \caption{Simulation of the detection system deployment. The \emph{first row} displays the original signal (acceleration over time steps), the \emph{second row} shows the logistic regression model's output (fall event detection), and the \emph{third row} presents the CNN output (human fall detection).}\label{fig:res1}
\end{figure}

The deployment experiment replicates the intended real-time fall detection process as illustrated in Fig.~\ref{fig:res1}:  The signal (blue) undergoes a two-stage classification. First logistic regression for initial event detection (orange) and second STFT and CNN classification (red). A spike in the orange graph marks every time the logistic regression detected an event itself and therefore a STFT was calculated. A spike in the red line marks every time the CNN classified one of those STFTs as a human fall. The signal shown in Fig.~\ref{fig:res1}  indicates a period of 3 hours in a room. Only a rough indication of around mid-afternoon was provided as a label by employees of the nursing home. 
Similar to the laboratory data analysis, we processed 10-second sequences individually by moving a window through the data chronologically. For each window, we calculated an input vector for the logistic regression. The frequency of this calculation depended on the step size of the shifting window. In our experiment, the window size matched the step size, resulting in non-overlapping signals. With a signal length of 3 hours, this led to a total of 1080 windows.

\section{Results and Discussion}\label{sec:results}

\subsection{Augmentation Results}

\noindent The initial experiment focused on data augmentation through duplication, which served as a baseline for the augmentation methods. It was observed that the precision of the model, which was trained on data without augmentation, fluctuated around 30\%, regardless of the duplication level. The duplication augmentation involved replicating each entry in the training set by a specific value, with tested values ranging from 1 to 30. As duplication increased, models trained on augmented data showed improved precision. However, the precision plateaued beyond a duplication value of about 10, likely because of a nearly balanced dataset at this point. Despite this, a higher duplication level reduced the performance gap between the all-inclusive model and the baseline model. Further experiments for statistical smoothing were deemed necessary, but an overall improvement with data augmentation was clear, with the highest precision recorded at 65.24\% for the model trained solely on augmented data at a duplication value of 20.

\begin{figure}[t]
  \centering
  \includegraphics[width=\textwidth]{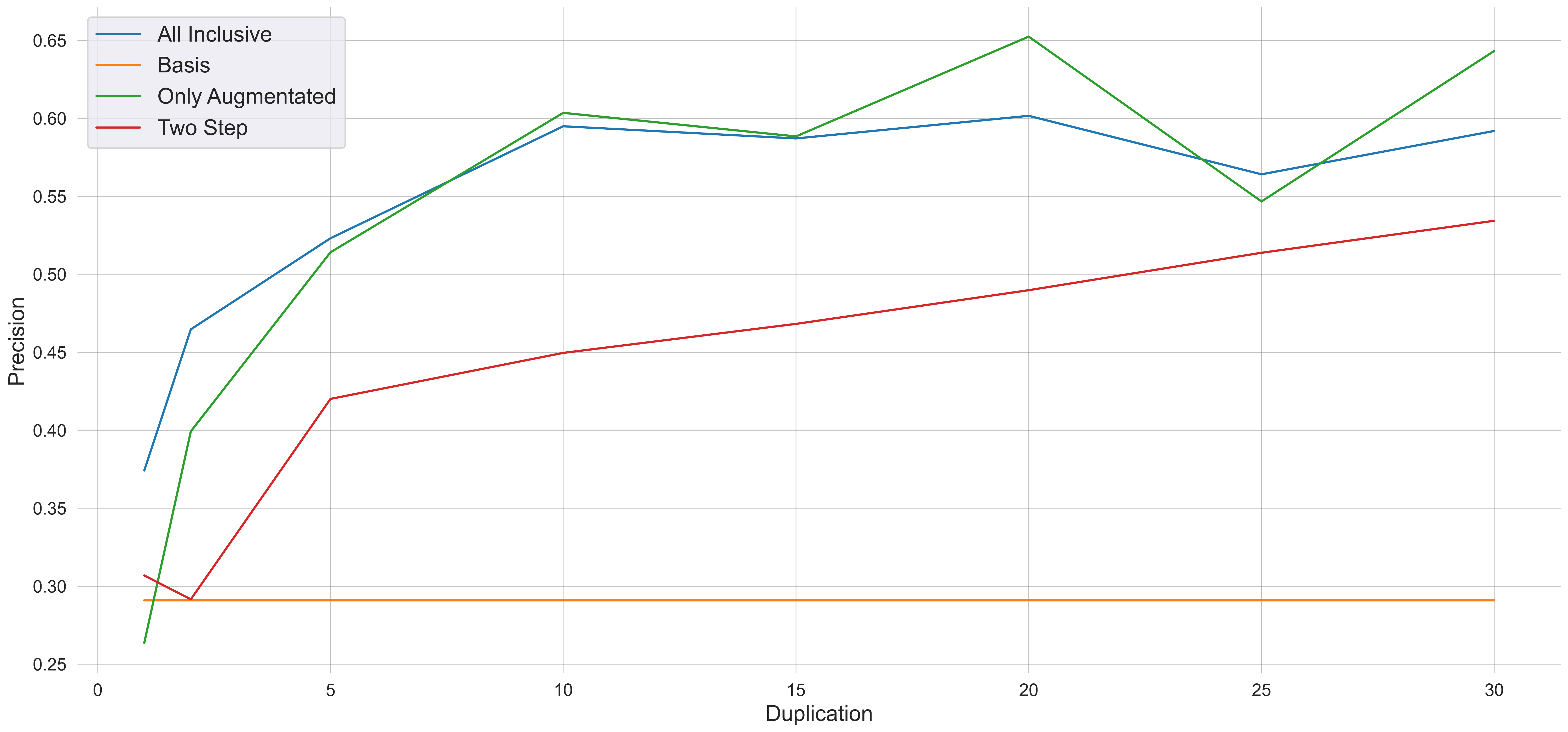}
  \caption{Performance of different training approaches with increasing amounts of duplicated dummy falls as an augmentation method. \emph{Green}: Training with only augmented data; \emph{Blue}: Training with both augmented and real data; \emph{Red}: Initial training with augmented data followed by fine-tuning on real data; Orange: Training without augmentation.}\label{fig:res3}
\end{figure}

\begin{figure}[t]
  \centering
  \includegraphics[width=\textwidth]{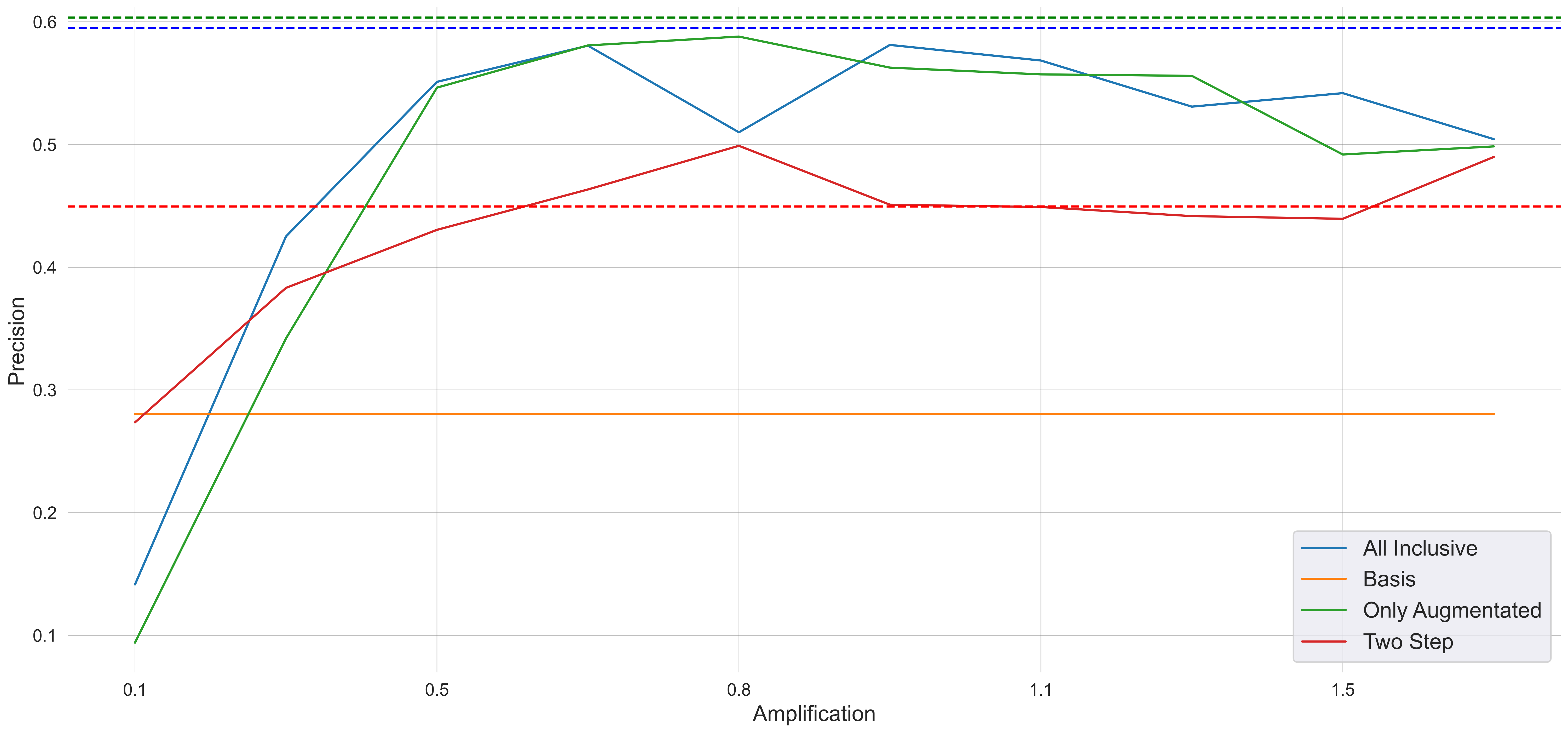}
  \caption{Performance of different training approaches with increasing amplification of dummy falls as an augmentation method. \emph{Green}: Training with only augmented data; \emph{Blue}: Training with both augmented and real data; \emph{Red}: Initial training with augmented data followed by fine-tuning on real data; \emph{Orange}: Training without augmentation. \emph{Dotted lines} indicate performance with 10- times duplication as the augmentation method..}\label{fig:res4}
\end{figure}

Following duplication, amplification was employed as another method of data augmentation. As described in Sec.~\ref{sec:augmentation}, this involved isolating events and adjusting their signal amplitude, with the augmented dataset being adjusted to balance dummy and non-dummy data, as informed by the findings from the duplication experiment. A reduction in precision was observed in the all-inclusive and only augmented models when signal amplitude was decreased, indicating effective isolation of the signal during the augmentation process. Optimal performance for the augmented models was found within an amplification range of 0.7 to 1.3. This suggested that amplification might not be as significant as the addition of similar data variants. Precision values were compared using a duplication of factor 10 for dataset size comparability as seen in Fig.~\ref{fig:res4}.

In the final comparison of augmentation methods, each experiment's baseline model was contrasted with its all-inclusive counterpart.
The primary focus was on the change in precision between the unaugmented and all-inclusive models, rather than on their absolute precision. 
The trend was unmistakable: as the dataset became more balanced with a consistent presence of unchanged dummy data, precision improved with the use of data augmentation, underscoring the importance of data balance in achieving model accuracy.

\begin{figure}[t]
  \centering
  \includegraphics[width=\textwidth]{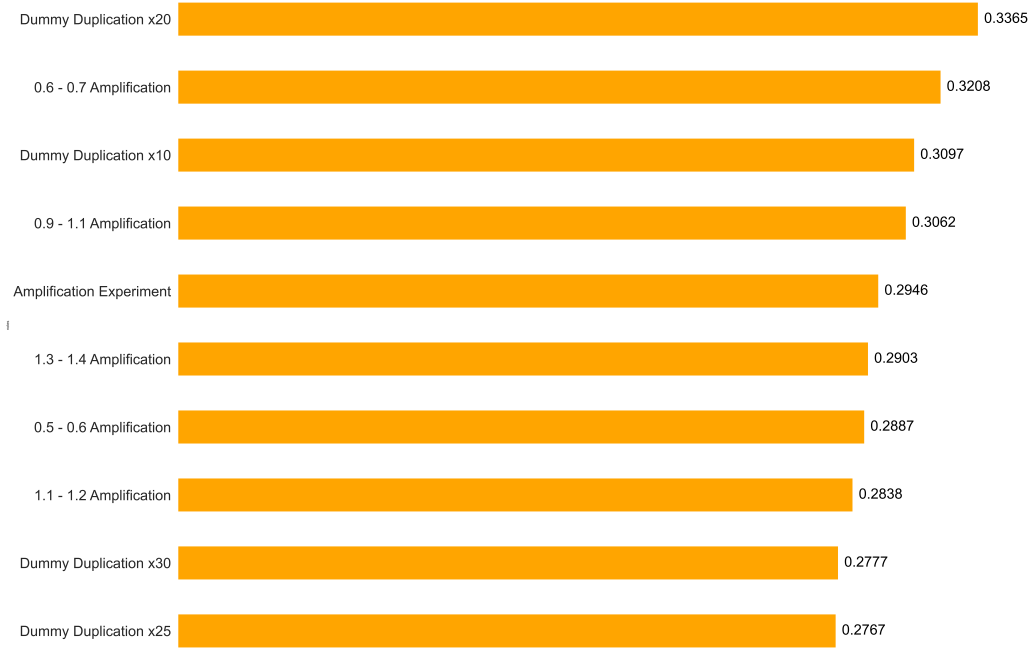}
  \caption{ Comparison of different augmentation methods. Duplications and small amplifications yield the best results.
  The \emph{orange bars} represent the model's precision after applying the respective augmentation methods (denoted on the y-axis).
  }\label{fig:res5}
\end{figure}

The experiments hint an inherent limitation of the underlying dataset in diversity and volume, which present a breeding ground for potential overfitting within the CNN models. If the dataset fails to encapsulate the rich spectrum of variations and nuances which are inherent in real-world scenarios, the models' ability to generalize beyond the provided data and not overfit is restricted. Additionally, data augmentation methods which enhance the model's ability to generalize cannot be fully tested if the validation is also too limited.

Moreover, the effectiveness of duplication as an augmentation technique, while demonstrating notable performance improvements, also raises concerns. The success of this method may indicate that the model tends to focus on certain cases in the data, which could affect its ability to generalize to new, unseen cases. In addition, the minimum change required during amplification (about 1) to achieve optimal performance also indicates the risk of limited diversity in the amplified signals. The closer the augmented signals are to the signals of the training set the better the model performance. This could be due to the similarity between test and training data independently of the fold. Experiments done with VAEs for date augmentation have shown restricted variance within the latent space, which further supports these indications. This restriction implies clustering of points, suggesting a lack of diverse representations within the dataset and potentially hindering the model's ability to capture a wide range of characteristics essential for robust generalization. Factors inherent in the dataset, such as the consistent representation of dummy falls and uniform fall motions directed straight to the ground, likely contribute to this limited variance. Such uniformity could limit the model's exposure to different fall patterns, which could increase its tendency to overfit. However, the challenge in comparing augmentation methods arises from the significant similarity between test and training data. This similarity makes it difficult to robustly assess the generalization abilities of the model and makes it difficult to identify possible overfitting tendencies. 

\subsection{Deployment Results}

\noindent The logistic regression identified 39 events within these windows, and the CNN pinpointed one of these events as a human fall. 

\begin{figure}[t]
  \centering
  \includegraphics[width=\textwidth]{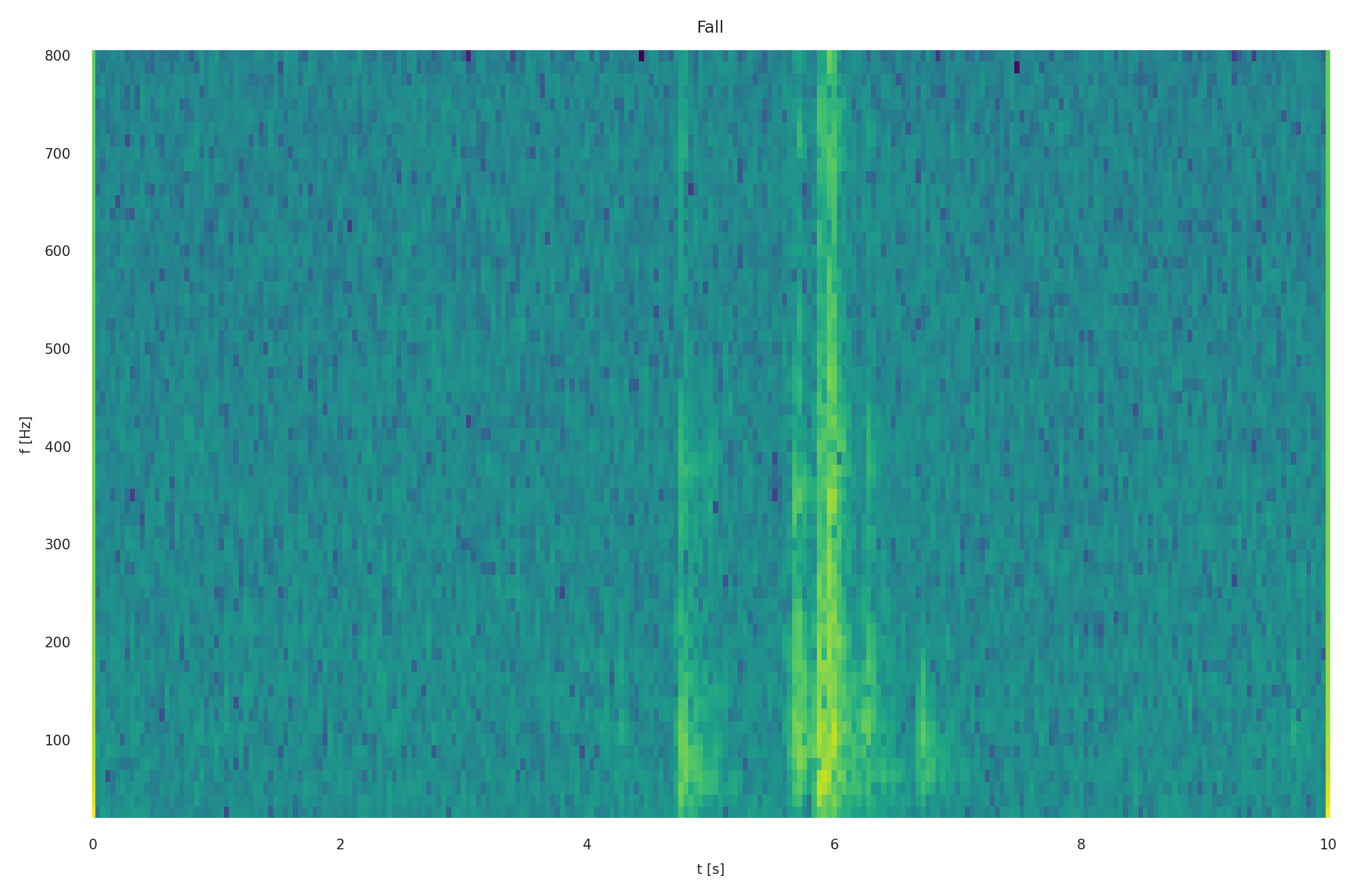}
  \caption{Spectrogram visualizing the detected fall event}\label{fig:res2}
\end{figure}

Figure~\ref{fig:res2}  displays the corresponding spectrogram of this fall. Currently, validating real falls is not feasible due to imprecise time data and various environmental vibrations in actual measurements. Although our model detects potential falls, we cannot verify the accuracy of these predictions. Moreover, establishing a direct link between Lab Data and Real Data remains elusive. To address this uncertainty, we recommend rigorous testing of the model in a real-world setting post-implementation of necessary measures. While the model effectively distinguishes falls from noise in artificial scenarios, its performance in real environments remains unproven. However, the model demonstrates sensitivity in discerning fall events without triggering frequent false alarms. Even if not all falls are detected during test deployments, individual event identifications can serve as valuable data for refining the model further.

\section{Summary and Outlook}

\noindent The approach of utilizing vibration-based fall detection systems presents several benefits.
Firstly, it offers a high level of reliability in detecting falls accurately, as it doesn't rely on manual activation like user-activated alarms or community alarm systems.
This eliminates the risk of the alarm button being unreachable during an emergency.
Additionally, it overcomes the issue of patients forgetting to wear or removing wearable devices, which commonly occurs with automatic wearable fall detectors.
Moreover, vibration-based systems are minimally invasive, thus preserving privacy better compared to camera-based alternatives.
They function independently of the patient's actions, reducing the burden on the user and ensuring continuous monitoring, even in private settings like bathrooms where falls are prevalent. Overall, vibration-based fall detection systems offer a balance of reliability, privacy preservation and user comfort, making them a promising solution for addressing fall-related risks in various settings, including nursing homes and home care environments. 
This study identified several challenges in developing this type of desirable monitoring system for nursing homes.

A crucial issue identified is the missing variance in lab data, hindering the application of sophisticated augmentation methods and the development of unbiased models.
To support ongoing research efforts in this area, the comprehensive dataset compiled will be disseminated as an integral part of this investigation.
Of particular note is the potential for synergistic merging of this dataset with other datasets in the field of fall detection, which promises to improve data fidelity and thereby advance the scientific study of the critical area of fall detection in nursing homes.
With only 66 instances of dummy case events and a mere 22 recorded falls, the dataset lacks both in volume and diversity. 
This limitation raises concerns about potential bias, particularly towards nursing home residents with a weight of 75~kg, neglecting the variability in fall characteristics across different weight categories. 

To mitigate this issue, we propose the generation of additional dummy data with a deliberate focus on introducing variability. Potential strategies include adjusting dummy mass through the incorporation of additional weights or modifying the distance between sensors and falling points in incremental steps. Additional data collection in real nursing home settings, possibly using a secondary device like dedicated fall detection wearables, might be of great interest as well. Enriching the dataset with diverse variations holds the promise of fostering a more robust model, enhancing its generalization capabilities, and preparing it for potential future integration of more sophisticated augmentation techniques.

Although the two-stage AI approach, which combines a simple logistic regression model and a relatively small CNN, was successful, more challenging real-world data might require a more complex setup with many hyperparameters. Therefore, more sophisticated hyperparameter tuning algorithms are necessary~\citep{bart21i}. Furthermore, online-machine learning approaches could be beneficial to adapt the model to the real-world data~\citep{mont20a,bart23c}.

Further research into improving sensors using AI is being carried out as part of the ShapeFuture project. This project is focusing primarily on improving the robustness of models based on sensor data and addressing the imbalance of sensor data. These problems have already been identified as critical in the course of this case study. 

In summary, the two-step approach utilizing the logistic regression demonstrates effective discrimination between events and noise. Furthermore, enhancing the performance of the CNN through techniques such as hyperparameter tuning and data augmentation proves significantly beneficial, surpassing the initial baseline performance. These findings underscore the potential of the proposed model architecture, emphasizing the need for ongoing refinement and validation in real-world environments.

\section*{Acknowledgements}

\noindent This research was supported by the BMWI (ZIM) project focused on developing sensor technology and AI algorithms for fall detection in nursing bed environments. Sensor technologies enhanced by AI will be further developed in the ShapeFuture project aimed at ensuring European ECS value chain sovereignty through advancements in automotive applications. We extend our gratitude to all partners, researchers, and institutions involved for their invaluable contributions and continuous support, significantly advancing our work in fall detection systems.
  
\appendix
\bibliographystyle{elsarticle-harv} 
\bibliography{bart24n}

\end{document}